\documentclass[5p,times,twocolumn]{elsarticle}

\usepackage{graphicx} 
\usepackage{booktabs} 
\usepackage{url}
\usepackage{tabularx}
\usepackage[T1]{fontenc}
\usepackage{lmodern}
\usepackage{multirow}
\usepackage{appendix}
\usepackage{hyperref}

\begin{document}

\begin{frontmatter}



\title{Advancing Reliable Synthetic Video Detection: Insights from the SAFE Challenge
}

\author{
Kirill Trapeznikov\textsuperscript{1},
Gabriel Mancino-Ball\textsuperscript{1},
Jonathan Li\textsuperscript{1},
Paul Cummer\textsuperscript{1},
Jai Aslam\textsuperscript{1},
Danial Samadi Vahdati\textsuperscript{2},
Tai Nguyen\textsuperscript{2},
Matthew C. Stamm\textsuperscript{2},
Peter Bautista\textsuperscript{3},
Michael Davinroy\textsuperscript{3},
Laura Cassani\textsuperscript{3},
Jill Crisman\textsuperscript{4}
}

\address{
\textsuperscript{1}\{kirill.trapeznikov, gabriel.mancino.ball, jonathan.li, paul.cummer, jai.aslam\}@str.us, STR; \\
\textsuperscript{2}\{ds3729, tdn47, mcs382\}@drexel.edu, Drexel University; \\
\textsuperscript{3}\{pbautista, mdavinroy, lcassani\}@aptima.com, Aptima, Inc.; \\
\textsuperscript{4}jill.crisman@ul.org, UL Research Institutes
\vspace{-10ex}
}

\begin{abstract}
The proliferation of generative video technologies has intensified the need for reliable methods to detect and characterize synthetic media. To address this challenge, we organized the \href{https://safe-video-2025.dsri.org}{SAFE: Synthetic Video Detection Challenge}, co-located with the \textit{Authenticity and Provenance in the Age of Generative AI (APAI) Workshop }at ICCV 2025.  The competition invited participants to develop and evaluate algorithms capable of distinguishing real from synthetic videos under fully blind evaluation conditions with over 600 submissions from 12 teams over a 90 day span. Hosted on the Hugging Face platform, the challenge comprised two primary tasks: (1) detection of synthetic video content generated by diverse state-of-the-art models, and (2) detection of synthetic content following common post-processing operations such as resizing, re-compression, motion blur and others. The challenge data consisted of 13 modern high quality synthetic video models with generated content matched to real videos from 21 diverse and challenge sources, all adding up to 20 hours of 6,000 video samples. This paper describes the challenge design, dataset construction, evaluation methodology, and outcomes, offering insights into the generalization and robustness of contemporary synthetic video detection methods. Our findings highlight measurable progress in cross-generator generalization but also persistent vulnerabilities to post-processing artifacts. \url{https://safe-video-2025.dsri.org}

\end{abstract}





\end{frontmatter}

\section{Introduction}

The emergence of large-scale generative video models spanning diffusion, transformer, and adversarial architectures, has rapidly blurred the boundary between authentic and synthetic media. Modern text-to-video and image-to-video systems can synthesize human motion, environmental dynamics, and complex scenes with heightened realism, creating both opportunities for creativity and new challenges for verifying media authenticity. As generative systems evolve, they increasingly produce content that defeats traditional semantic inconsistency markers.

Ensuring video authenticity and provenance has therefore become a central problem across journalism, digital safety, fraud prevention, and scientific integrity. Detecting synthetically generated or manipulated videos requires analytics that are robust to visual variability, resilient to post-processing, and generalizable across unseen generation techniques. However, most existing benchmarks provide fixed and openly available datasets, which makes it difficult to keep pace with the rapidly changing generative landscape. These datasets are also typically limited to facial or human-centric content, leaving open questions about robustness across broader semantic domains.

The SAFE: Synthetic Video Detection Challenge was designed to fill this gap. Executed using Hugging Face infrastructure, SAFE introduced a fully blind, model-submission-based framework where all evaluations were run against a sequestered dataset unavailable to participants. This approach ensured fairness and a closer approximation to real-world forensic workflows, where detectors must operate on unseen sources and cannot rely on metadata or file-format cues. The challenge received over 600 submissions from more than twelve teams over a 90-day period. Two tasks were released: \textit{(Task 1) Synthetic Video Detection:} baseline evaluation of generalization across modern high quality generators and diverse real-world content sources; and \textit{(Task 2) Post-Processed Video Detection:} assessment of robustness to realistic transformations such as resizing, compression and other changes common during online dissemination.

Together, these tasks provided a structured, multi-factor test of both core detection capability and operational robustness, offering a view into how current synthetic-video detectors perform under controlled yet realistic blind conditions. This paper makes the following key contributions: 
\begin{itemize}
    \item \textit{Evaluation on realistic synthetic and real content} The challenge evaluated detection performance (Task 1) across 13 modern high quality text+image-to-video (TI2V) and other models with generated content matched to videos from 21 diverse and challenging real sources. The data comprised 20 hours of video across 6,000 samples of content designed to match the distribution of videos found in the wild. 
    
     \item \textit{A dual-task design measuring generalization and robustness.} While Task 1 established baseline performance, Task 2 simultaneously evaluated post-processing degradations to reveal detector vulnerabilities. Task 2 data consisted of videos with 14 post-processing techniques commonly applied to videos when shared online (i.e. resizing, recompression, etc.)
     
    \item  \textit{Fully blind benchmark for synthetic video detection.} Unlike prior open-dataset efforts, SAFE required participants to submit their models to be evaluated on sequestered data using the same computational platform. Reproducing real world settings, participants had no knowledge of video content and sourcing. The competition also reduced the risk of over-training by restricting the number of daily submissions and providing  metrics only on the public subset of the data.


    \item \textit{Comprehensive analysis of state-of-the-art detectors.} This paper reports aggregated findings from top submissions, identifying several key trends such as (i) strong detection performance of unmodified synthetic video content in the fully blind competition setup (ii) non-trivial detection degradation after common post-processing, (iii) emerging good practice for detector designs such as using large pre-trained vision backbones and autoencoding augmentation (encoding input data into a latent representation and subsequently reconstructing the original input via a decoder) and (iv) surprising transferability of detectors to novel synthetic sources. 
\end{itemize}
\section{Background and Related Work}
Early manipulated media benchmarks primarily targeted faces and controlled capture environments. FaceForensics++ introduced one of the first large-scale benchmarks for facial manipulation detection, encompassing four automated generation methods (Face2Face, FaceSwap, DeepFakes, and NeuralTextures). The study reported pronounced performance degradation under compression, underscoring the sensitivity of detectors to video quality and encoding conditions \cite{rossler2019faceforensicspp}. Building on this foundation, the DeepFake Detection Challenge (DFDC) dramatically expanded scale and diversity, releasing over 100{,}000 videos featuring 3{,}426 paid actors and multiple pipelines to facilitate community-wide benchmarking \cite{dolhansky2020dfdc}.

Subsequent efforts have focused on improving robustness and representativeness. DeeperForensics-1.0 introduced 50k real and 10k forged videos accompanied by extensive ``in-the-wild'' perturbations, providing a framework for evaluating models under real-world noise and transformations \cite{jiang2020deeperforensics}. Broader demographic coverage was addressed by KoDF, a large-scale dataset focused on Korean subjects with 175k fake and 62k real clips generated by six synthesis models, highlighting distribution shifts relative to prior corpora \cite{kwon2021kodf}. Moving beyond highly curated settings, the WildDeepfake dataset collected 7k face sequences from 707 deepfake videos gathered entirely from the internet; baseline detectors trained on curated datasets degrade substantially on this challenging distribution \cite{zi2021wilddeepfake}.

As image-to-video and text-to-video generators matured, recent work broadened scope beyond face-centric manipulations to general (non-face) synthetic video. The GenVideo benchmark reports large-scale real and generated videos spanning diverse categories and generation techniques to study cross-source generalization in AI-generated video detection \cite{chen2024demamba}. DF40 aggregates content from 40 distinct deepfake techniques to facilitate next-generation cross-technique evaluation at scale \cite{yan2024df40}. Complementary evaluations such as VANE-Bench probe large multimodal models on anomaly and inconsistency detection (including synthetic videos), illuminating sensitivity to subtle generative artifacts even when the task is framed as question answering rather than binary forensics \cite{bharadwaj2024vane}.

Recent datasets have further emphasized open-set, multilingual, and multimodal evaluation. Deepfake-Eval-2024 curates deepfakes from 88 websites across 52 languages, demonstrating substantial performance drops for existing detectors when tested on contemporary content compared to prior academic benchmarks \cite{chandra2025deepfakeeval}. SocialDF focuses on social-media contexts, compiling 2k real and synthetic videos and proposing multimodal detection approaches aligned with user-generated media verification \cite{batra2025socialdf}. MAVOS-DD extends this trend to multilingual, audio–visual deepfake detection, covering 8 languages and over 250 hours of content; results show persistent challenges in generalizing to unseen languages and generator architectures under open-world conditions \cite{croitoru2025mavosdd}.

Another important evaluation dimension is robustness to post-processing. Videos disseminated online are often compressed, resized, re-encoded, or filtered, masking the forensic traces that detectors rely on. The FaceForensics++ authors specifically highlighted degradation in performance under compression and lower-quality settings \cite{rossler2019faceforensicspp}. 

Community analyses around the DFDC effort (e.g., the Partnership on AI report) highlighted deployment-relevant considerations that differ from purely academic settings, motivating evaluations that better reflect operational constraints and real content flows \cite{pai2020dfdc}. Collectively, these works motivate the design decisions in SAFE: a blind, model-submission evaluation on hidden data with standardized compute, and task definitions that explicitly measure generalization to unseen generators (Task~1) and robustness to common post-processing operations (Task~2).

Taken together, these findings suggest 3 major challenges: (i) \textit{generalization}: detectors must work across unknown generators and domains, (ii) \textit{robustness}: detectors must tolerate common post-processing and real-world distribution shifts, and (iii) \textit{evaluation realism}: existing benchmarks often release fixed training and test sets that may be overfitted by the research community, reducing their real-world relevance. These challenges motivated the design of our competition.

\section{Competition Overview}
The SAFE: Synthetic Video Detection Challenge was hosted on the Hugging Face Hub. Running for four months, the competition ended at a media forensic conference at a major computer vision conference. Its overarching goal was to benchmark and advance methods for detecting AI-generated and manipulated video content under realistic, blind evaluation conditions. The challenge focused on two central aspects of video forensic performance: the ability to generalize across unseen generation models and domains, and the robustness of detection methods to real-world post-processing operations such as compression and resizing. All evaluations were conducted through containerized submissions executed in the same, offline compute environments to ensure fairness and reproducibility.
\subsection{Tasks} 
Our challenge consisted of two synthetic video detection tasks with increasing difficulty.

\textbf{Task 1 -- Detection of Synthetic Video Content} focused on identifying video clips generated from various high performing video generation models. No post processing operations were performed on the synthetic video clips. As for the real video clips, audio was stripped and each clip was split into 5-10 second chunks to mirror the synthetically generated clips. A comprehensive list of the 21 video generation sources and 13 video generation models is provided in Tables~\ref{tab:real_video_sources} and~\ref{tab:synthetic_video_sources}.  

\textbf{Task 2 -- Detection of Synthetic Video Content after Post-Processing} assessed the ability to detect synthetic content from post-processed video clips. Fourteen commonly used post processing techniques include upscaling, downscaling, motion blur, etc. A detailed list of the post processing techniques is provided in Table~\ref{tab:post_processing_technique}. 

\subsection{Evaluation Criteria}
The main metric for the competition was area under curve (AUC), balanced accuracy (BAC), defined as an average of the true positive (TPR) and true negative rates (TNR). The competition maintained a public and private leaderboard for each task where participants had access only to the public leaderboard for the duration of the competition. The public leaderboard provided a breakdown of the evaluation metrics across the entire public dataset as well as across each real and generated source. The names of each source were anonymized on the public leaderboard. This limited feedback and visibility allowed us to maintain a black-box testing scenario. The final evaluation for each task was based on BAC over the private dataset.

The dataset for each task consisted of a public and private set where the public set was a subset of the private dataset. The public dataset consisted of 10 out of 21 real sources and 6 out of 13 generators. The private set contained the entire dataset, including sources which were withheld from the public set. This design choice was made to incentivize participants to develop detectors that did not overfit on the public dataset.

\subsection{Setup and Rules}
The SAFE: Video Detection Challenge was a script-based, fully-blind competition format where no data was released for the duration of the competition. The competition was hosted through a custom version of the the Hugging Face competition framework \cite{hf_comp_2025}. Participants submitted their model by creating a private repository on huggingface.co with no restrictions on what their repo may contain. Participants could have included any code, packages, model weights, or environment within their repository. The only requirement was that participants had to submit a script file that read in our competition dataset and outputted a submission file in a specific format. Participants submitted their model by logging into our competition and authorized our framework to access their model. Our testing framework then evaluated participants' models with the following procedure as shown in Figure \ref{fig:comp_setup}: (i) environment is installed, (ii) network access is disabled, (iii) submission model and dataset are pulled and (iv) the evaluation script is executed and the results submission file is uploaded and evaluated.

During the evaluation step, we sequestered the participants' models from the Internet to prevent ex-filtration of the competition data or their potential use of externally hosted services. To level the field, all submissions were executed on the same compute resources of an NVIDIA L40S GPU with 48GB VRAM, 62GB RAM and 8 CPUs. All submissions were limited to 20,000 seconds and participating teams were limited to three submissions per day. Submission logs were automatically analyzed with the Qwen3 480 billion parameter coder model~\cite{yang2025qwen3} to provide teams with feedback related to errors during their submission process. Since we did not release any data during the competition, we provided several pointers to publicly available open source synthetic datasets to assist participants.

To further simulate real world settings, we required models to make a binary decision of either "real" or "generated" in their submission file. Therefore, the evaluation was performed at a confidence threshold chosen by each participant in each test sample. However, to facilitate further analysis, the submission file also had to contain a decision score and an average inference time for each test data point input file to allow us to compute AUCs.

\begin{table*}[tbh!]
\centering
\small
\setlength{\tabcolsep}{4pt}
\renewcommand{\arraystretch}{1.15}
\begin{tabularx}{\textwidth}{@{}l r r r >{\raggedright\arraybackslash}X l l@{}}
\toprule
\textbf{Source} & \textbf{Dur} & \textbf{Res} & \textbf{FPS} &
\textbf{Description} & \textbf{Enc} & \textbf{Split} \\
\midrule
vsp-trail-cam & 7.27 & 0.23 & 30.00 &
Wildlife trail cam & H.264 & public \\
vsp-drone & 7.37 & 0.23 & 29.89 &
Aerial drone footage & H.264 & public \\
vlog & 9.78 & 6.95 & 26.52 &
Vlog content & H.264 & public \\
time-lapse & 9.12 & 3.99 & 27.05 &
Time-lapse footage & H.264 & public \\
social-media-text-overlays & 7.47 & 0.90 & 28.96 &
Shortform videos with subtitles & H.264 & public \\
social-media-dance & 7.51 & 2.07 & 30.00 &
Short form dance videos & H.264 & public \\
social-media-composite-video & 7.07 & 0.89 & 30.00 &
Social media content & H.264 & public \\
lecture & 7.54 & 0.23 & 30.00 &
Academic lecture recordings & H.264 & public \\
drone & 7.11 & 2.02 & 30.00 &
Aerial drone footage & H.264 & public \\
documentary-2 & 8.66 & 4.54 & 30.29 &
Documentary clips & H.264 & public \\
documentary-1 & 6.68 & 2.07 & 29.97 &
Documentary clips & H.264 & public \\
\midrule
vsp-tv-talk-show & 6.64 & 0.23 & 25.00 &
TV talk show clips & H.264 & private \\
vsp-travel & 6.96 & 1.76 & 25.00 &
Travel and tourism & H.264 & private \\
vsp-selfie & 6.96 & 0.21 & 29.87 &
Self-recorded short form content & H.264 & private \\
vsp-production & 7.60 & 0.23 & 25.00 &
Broadcast footage & H.264 & private \\
sports & 7.07 & 4.00 & 30.94 &
Sports highlights & H.264, MJPEG & private \\
social-media-phone-camera & 7.17 & 0.88 & 29.34 &
Short form content & H.264 & private \\
internet-archive-sports & 7.22 & 0.11 & 26.92 &
Sports highlights & H.264 & private \\
internet-archive-pre-digital & 7.32 & 1.28 & 25.00 &
Pre-digital age footage & H.264 & private \\
dashcam & 7.60 & 2.16 & 10.00 &
Car dashcam footage & MPEG-4 & private \\
academic-selfie & 7.66 & 2.07 & 50.00 &
Talking head footage & H.264 & private \\
\bottomrule
\end{tabularx}
\caption{Real Video Sources in the challenge split by public and private sets. Dur(ation) is average duration of video in seconds. Res(olution) is average resolution in megapixels.}
\label{tab:real_video_sources}

\end{table*}
\begin{table*}[tbh!]
\centering
\small
\setlength{\tabcolsep}{4pt}
\renewcommand{\arraystretch}{1.15}
\begin{tabularx}{\textwidth}{@{}l r r r >{\raggedright\arraybackslash}X l c l@{}}
\toprule
\textbf{Source} & \textbf{Dur} & \textbf{Res} & \textbf{FPS} &
\textbf{Description} & \textbf{Enc} & \textbf{OS} & \textbf{Split} \\
\midrule
wan-2\_1-i2v-480p~\cite{wan2025} & 5.06 & 0.43 & 16 &
Open-source image-to-video model & H.264 & Yes & public \\
veo-2~\cite{veo2_2024} & 5.00 & 0.92 & 24 &
High-quality cinematic generation & H.264 & Yes & public \\
pixverse-v4\_5~\cite{pixverse_2025} & 5.37 & 0.61 & 30 &
General-purpose video generator & H.264 & No & public \\
kling-v2\_0~\cite{kling_2025} & 5.04 & 0.92 & 24 &
Commercial text-to-video model & H.264 & No & public \\
hunyuan~\cite{kong2024hunyuanvideo} & 5.37 & 0.86 & 30 &
Open-source high-resolution generator & H.264 & Yes & public \\
framepack~\cite{zhang2025framepack} & 5.01 & 0.92 & 30 &
Lightweight long-video generation framework & H.264 & Yes & public \\
\midrule
vidu-2.0~\cite{bao2024viduhighlyconsistentdynamic} & 4.03 & 0.92 & 32 &
Stylized and animated video synthesis & H.264 & No & private \\
video-01~\cite{minimaxvideo_2024} & 5.64 & 0.90 & 25 &
Photorealistic cinematic animation & H.264 & No & private \\
seedance-1-pro~\cite{gao2025seedance10exploringboundaries} & 5.04 & 2.09 & 24 &
High-resolution multi-shot generation & H.264 & No & private \\
runway\_gen4\_turbo~\cite{runwaygen4_2025} & 10.08 & 0.92 & 24 &
Professional cinematic video generation & H.264 & No & private \\
physicalai~\cite{Tang25AICity25}~\cite{Wang24AICity24}~\cite{Wang24MCBLT} & 10.00 & 2.07 & 30 &
Multi-view synthetic human activity dataset & H.264 & Yes & private \\
hunyuan-avatar~\cite{hu2025HunyuanVideo-Avatar} & 5.15 & 0.62 & 25 &
Audio-driven human animation & H.264 & Yes & private \\
cosmos-drive-dreams~\cite{nvidia2025cosmosdrivedreams} & 5.04 & 0.90 & 24 &
Synthetic driving scene generation & H.264 & Yes & private \\
\bottomrule
\end{tabularx}
\caption{Synthetic Video Sources in the challenge split by public and private sets. Dur(ation) is average duration of video in seconds. Res(olution) is average resoltion in megapixels.}
\label{tab:synthetic_video_sources}
\end{table*}
\begin{table*}[tbh!]
\centering
\small
\renewcommand{\arraystretch}{1.15}
\setlength{\tabcolsep}{6pt}
\begin{tabularx}{\linewidth}{@{}l >{\raggedright\arraybackslash}X >{\raggedright\arraybackslash}X@{}}
\toprule
\textbf{Technique} & \textbf{Description} & \textbf{Parameters} \\
\midrule
border & Add padded black border & width = min video dim*0.1 \\
color correction & Adjusts colors using LUTs & \textemdash{} \\
compr. libaom-av1 & AV1 Compression & crf=10-50 \\
compr. libx264 & H.264 Compression & crf=10-40 \\
compr. libx265 & H.265 Compression & crf=10-40 \\
upscale & increase res by 75\%, center-crop to o.g. res & \textemdash{} \\
downscale & decrease resolution by 75\% & \textemdash{} \\
frame interpolation & increased fps using minterpolate algorithm & fps=[16,24,30,60,2*current fps] \\
gaussian blur & gaussian blur with sigma $\propto$ video res & ref sigma=3 ref res=1080p \\
motion blur & Temporally averaging consecutive frames & \textemdash{} \\
noise & Adds gaussian noise & target PSNR=36dB \\
real camera & Recording of video using webcam & \textemdash{} \\
speed adjustment & Alters playback speed & scale factor $\in$ [0.25, 2.0] \\
text overlay & Superimpose text captions onto video & font size=min video  dimension * 0.05 \\
\bottomrule
\end{tabularx}
\caption{Post-processing and augmentation techniques used in Task 2, descriptions and parameters (if any)}.
\label{tab:post_processing_technique}
\end{table*}

\begin{figure}[tbh!]
\centering
\includegraphics[width=.49\textwidth]{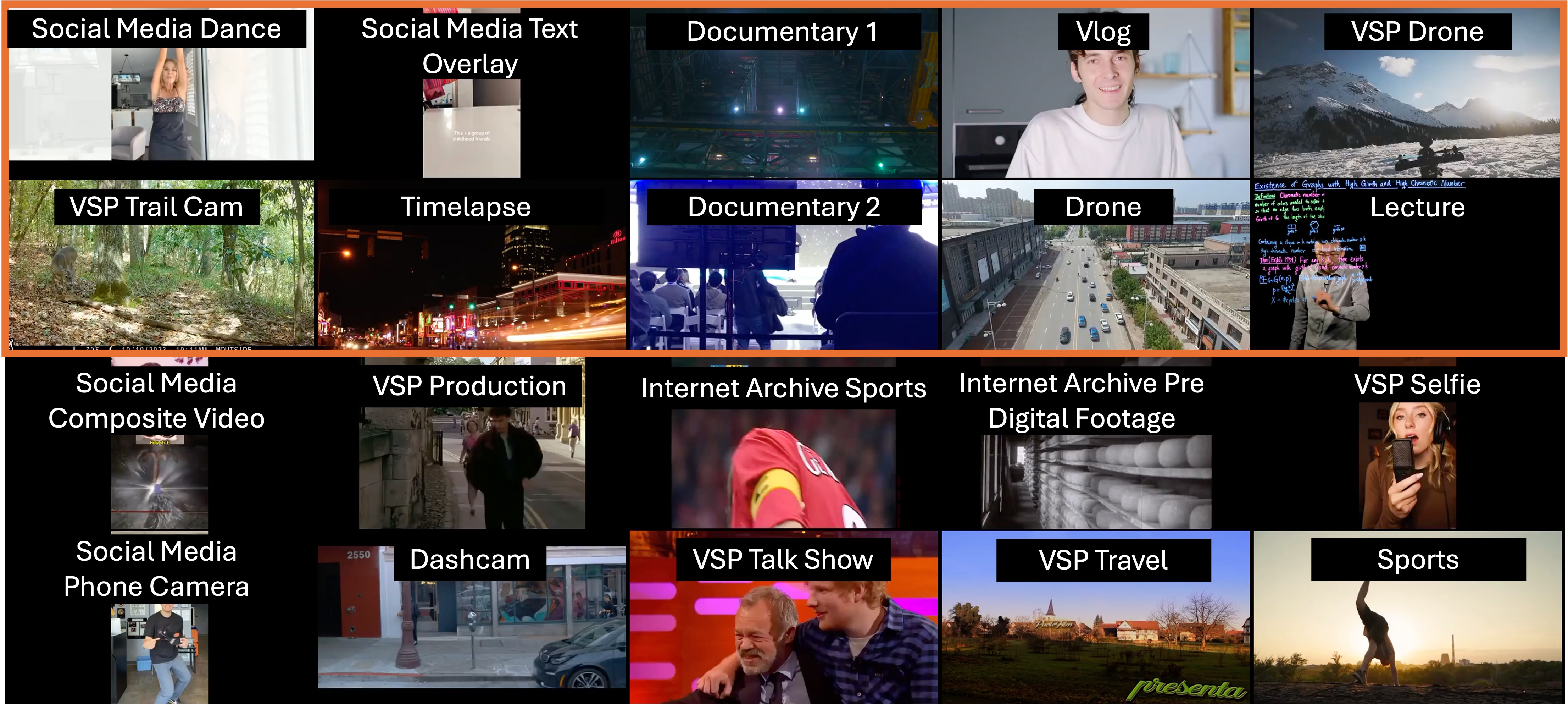}
\caption{Real Video Sources: a sample frame from each video source is shown. The top examples within the orange outline were in the public split and the remaining examples were in the private.}
\label{fig:real_sources}
\end{figure}

\begin{figure*}[tbh!]
\centering
\includegraphics[width=1\textwidth]{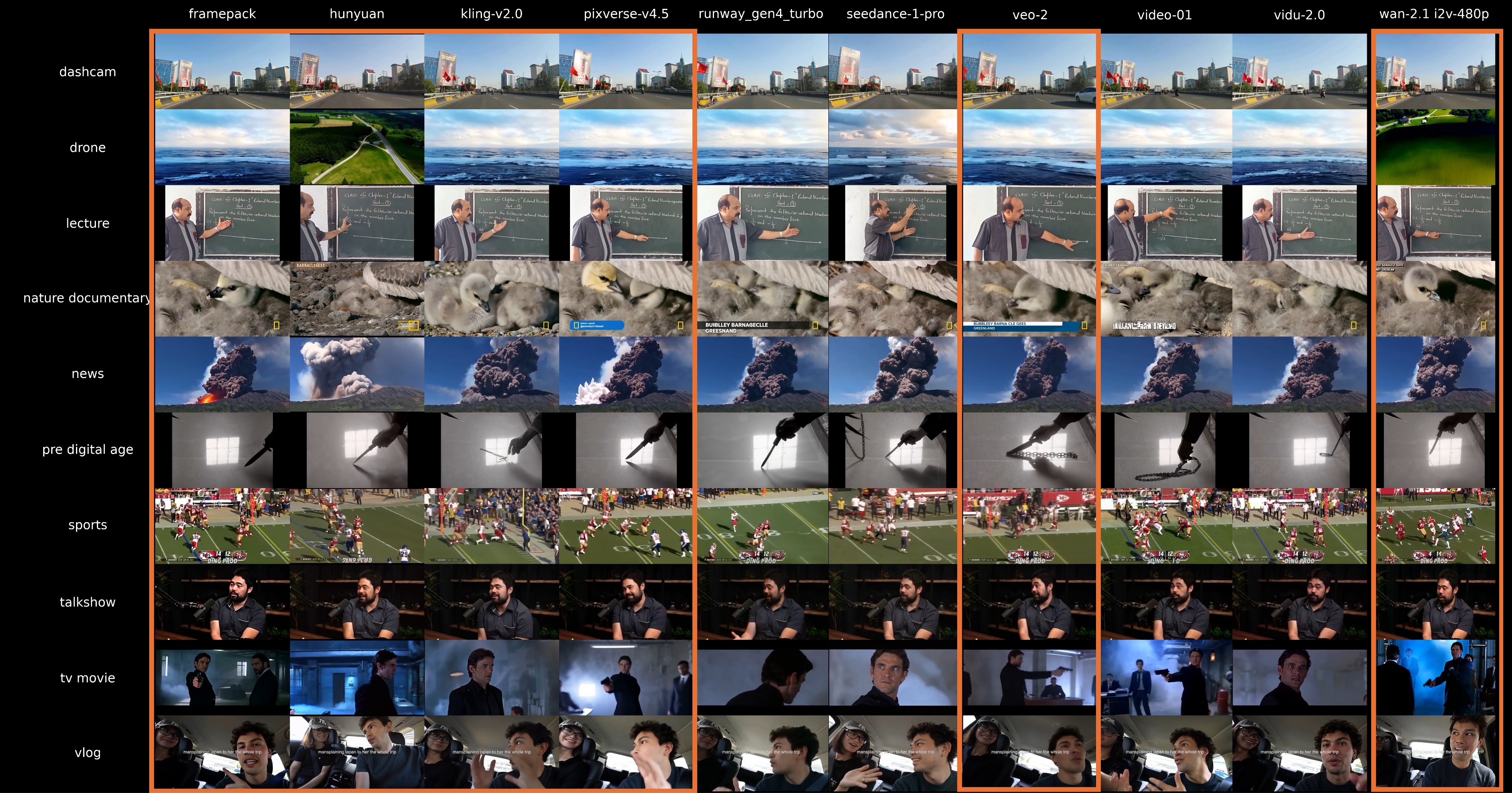}
\caption{Synthetic Video Sources: 10 different frame samples matching the content of the real videos for each TI2V generator (columns). The orange outlined sources are in the public split and the rest are in private.}
\label{fig:generated_sources}
\end{figure*}

\begin{figure}[tbh!]
\includegraphics[width=.48\textwidth]{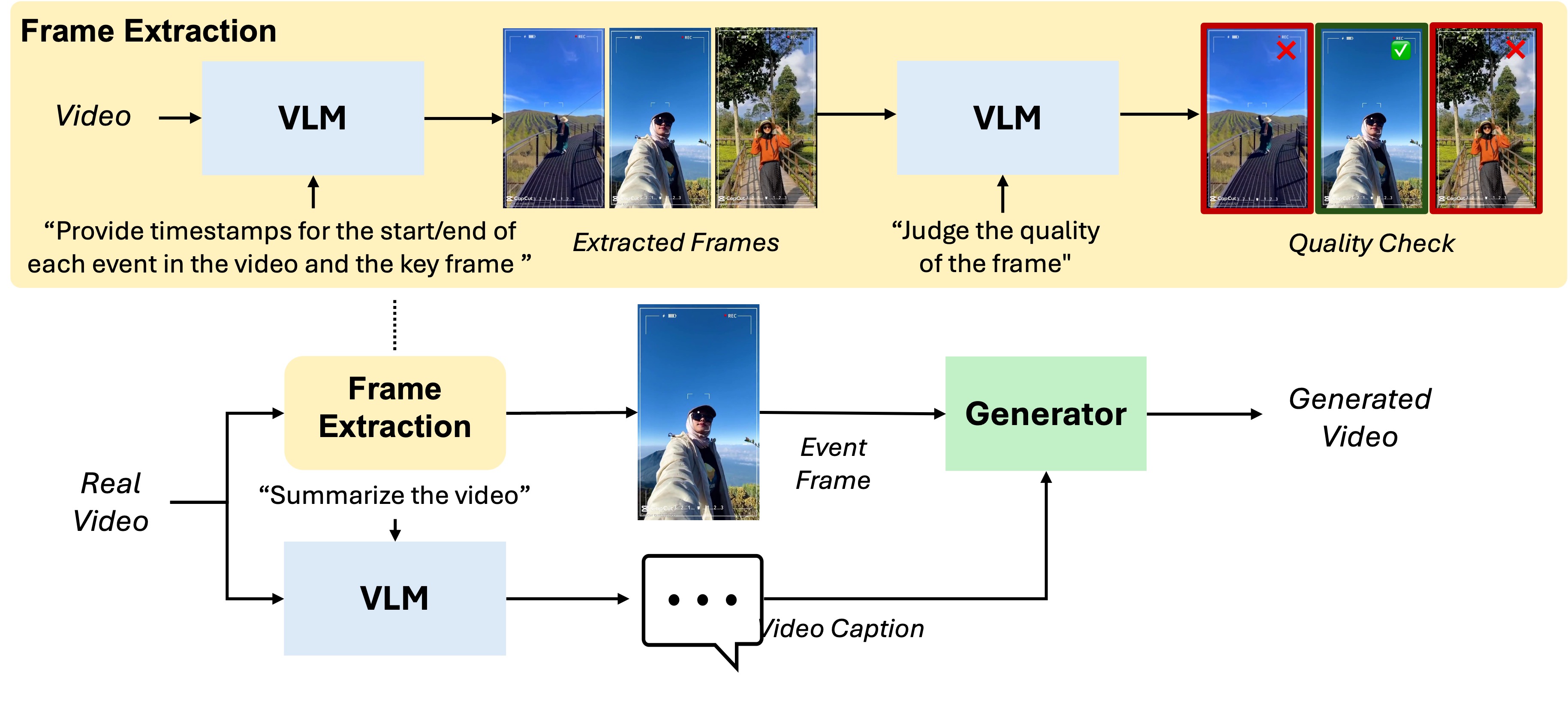}
\caption{Generation pipeline used for creating the dataset that matches real video content distribution.}
\label{fig:video_genertion_pipeline}
\end{figure}

\begin{figure}[tbh!]
\centering
\includegraphics[width=.48\textwidth]{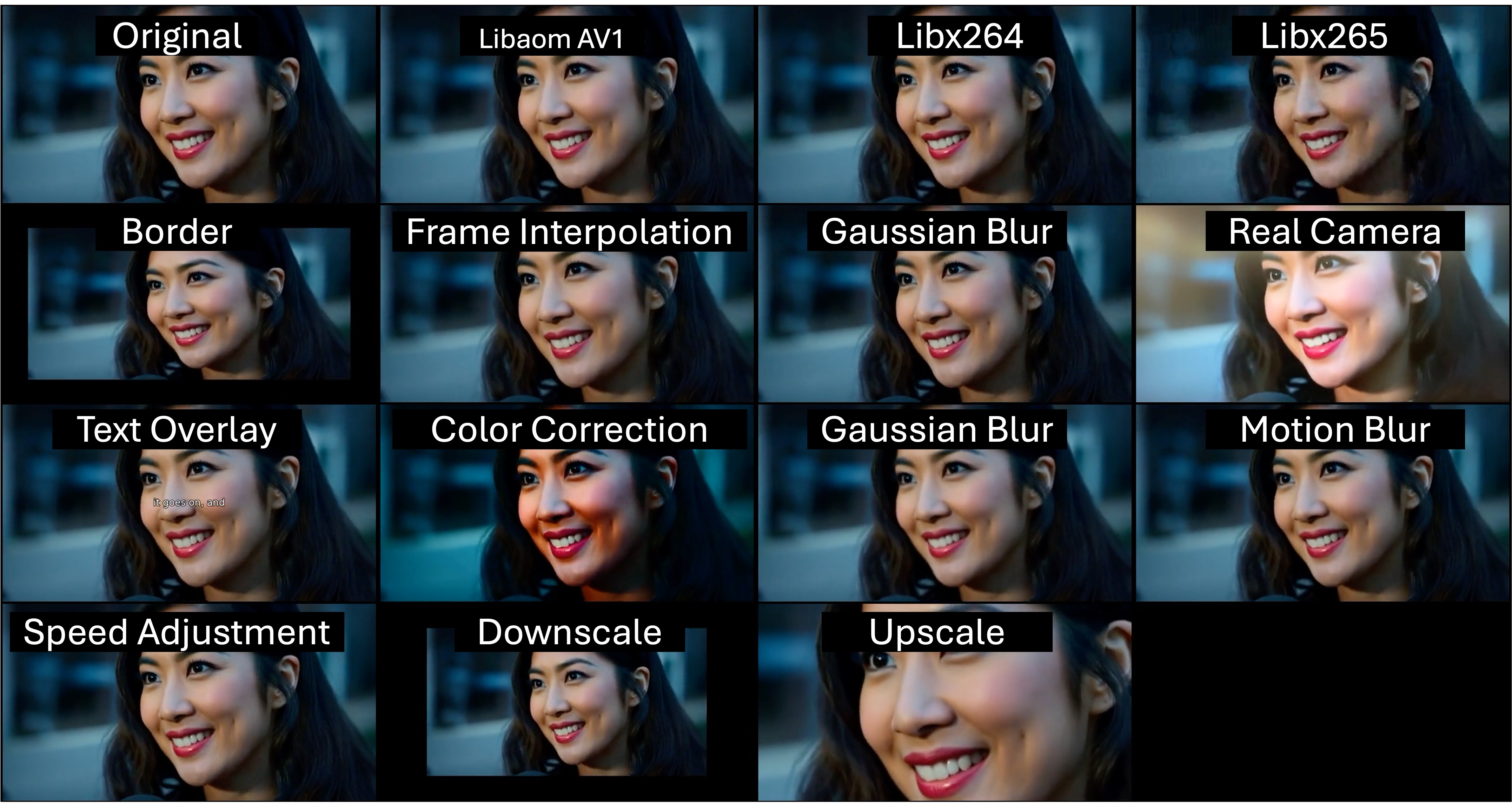}
\caption{A sample of frames showing the  visual quality per augmentation applied in Task 2}.
\label{fig:video_augs}
\end{figure}

\section{Dataset Description}
For the challenge, we constructed a comprehensive dataset consisting of 13 state of the art Text+Image-To-Video (TI2V) models and other generator models and wide range of real videos from 21 different sources, totaling 6,000 samples and 20 hours of video. We designed the synthetic data to be of high quality and representative of what people would likely encounter in the wild, while being semantically aligned with the real videos (i.e. have the same content). For Task 2, we reused a subset of Task 1 data to apply a common set of post-processing operations.
\subsection{Real Video Sourcing}
The real videos in  Task 1 of the challenge were curated from 21 diverse sources, encompassing a broad spectrum of content including recent videos such as those found on social media, drone footage, sports, time lapse videos,  smartphone footage, and shows of various production quality available on video sharing platforms (VSPs). We designed the dataset to be as representative of the real distribution videos found on the Internet, so we sourced from a wide variety of platforms such as VSP, social media, the Internet archive, etc. To make the data more challenging, we included more obscure content such as pre-digital age videos, trail cam footage, dashcam videos, and academic selfie data. We also included altered content such as video with text overlays and videos that were a composite of multiple shots (all very common on the Internet). The dataset comprised 1,064 real videos with approximately 50 videos per source. We partitioned the data such that 10 sources were allocated to the public data split, with the remaining 11 sources reserved for the private evaluation split. The video sources, description, and partition are listed in Table~\ref{tab:real_video_sources}. To ensure consistency with the synthetically generated videos, all real videos were processed by removing audio content and truncating to segments of 5-10 seconds in duration. The spatial resolutions of these videos varied considerably, ranging from 240p to 4K. Figure \ref{fig:real_sources} shows a sample of frames per each source with the orange outlined frames indicating the public set.
\subsection{Synthetic Video Generation}
In constructing the synthetic video data, we focused on (i) replicating commonly used video generation tools used in the wild by including high quality, readily available closed and open source models, and (ii) matching the content of the real videos by seeding the generation with a starting frame and descriptions derived from real content.

\textbf{Generators}: In Task 1, we utilized 13 state-of-the-art models, including 10 diffusion based TI2V, one Audio-Driven model (Hunyuan-Avatar), and two physical environment simulation models (PhysicalAI-Autonomous-Vehicle-Cosmos-Drive-Dreams, PhysicalAI-SmartSpaces). 100 image-text pairs served as inputs to each of the 10 generative models to create 1000 synthetically generated videos with a distribution of content similar to the real videos (see details on our video generation pipeline below). Out of the 10 model sources, 6 were used for the public dataset, and 4 were used in the private split. The private dataset was supplemented with 100 videos generated using audio driven Hunyuan-Avatar, 100 synthetic dash cam videos sampled from PhysicalAI-Autonomous-Vehicle-Cosmos-Drive-Dreams, and 100 synthetic indoor workspace videos sampled from PhysicalAI-SmartSpaces. Each video source used in the synthetically generated portion of the dataset is listed in Table~\ref{tab:synthetic_video_sources}. After generation, the videos were left as is with no modifications or post processing. The generated videos varied in duration from 5 to 10 seconds and in resolution from 360p to 4K. Figure \ref{fig:generated_sources} shows a sample of frames per each source with the orange outlined box indicating the public set.

\textbf{Video Generation Pipeline}: To align the content of synthetic and real videos, we developed an automated generation pipeline shown Figure~\ref{fig:video_genertion_pipeline}. TI2V models require a starting image and a text description of the desired content for video generation. To ensure that the distribution of content in generated videos matched that of real videos, we designed the following process. We uniformly sampled 100 videos across all categories from the real video dataset. For each video, we extracted a clear and semantically meaningful frame by instructing Qwen2.5-VL \cite{Qwen-VL} to identify event boundaries within the video, where an event is defined as a segment containing a unique action. For each identified event, Qwen2.5-VL selected one representative frame, then performed a quality assessment based on clarity and absence of artifacts, ultimately selecting the frame it deemed best matched the criteria as input for video generation. For the text prompt, we instructed Qwen2.5-VL to generate a detailed description of each video's content. This method produced 100 text-image pairs matching the category distribution of real videos, which served as inputs for the text/image-to-video models. By using identical text-image input pairs across all generators, we isolated generator performance from content variability, enabling more accurate comparative evaluation.
\begin{table}[tb]
\small
\centering
\begin{tabular}{ll}
\toprule
\textbf{Team Name} & \textbf{Organization} \\
\midrule
\textbf{GR}IP-UNINA & University of Naples Federico II \\
\textbf{IS}PL-Realynx & ISPL, Politecnico di Milano \\
\textbf{DA}SH & DASH Lab / Sungkyunkwan University \\
\textbf{LE}mma1727 & DASH Lab / Sungkyunkwan University \\
\textbf{TR}ustbees & Truebees \\
\textbf{SH}ahidmuneer & DASH Lab / Sungkyunkwan University \\
\textbf{DX} & DX Lab / Sungkyunkwan University \\
\textbf{BA}seline & MISL / Drexel University \\
\bottomrule
\end{tabular}
\caption{Team names and institutional affiliations for the top participating teams in the SAFE Video Detection Challenge.}
\label{tab:team_mapping}
\end{table}

\begin{table}[tb]
\centering
\small
\begin{tabular}{lcrrrrrr}
\toprule
\textbf{Metric} & \textbf{Task} & \textbf{GR} & \textbf{IS} & \textbf{LE} & \textbf{DA} & \textbf{TR} & \textit{BA} \\
\midrule
\multirow{2}{*}{\textbf{BAC}}
 & 1 & \textbf{0.86} & 0.85 & 0.82 & 0.84 & 0.77 & 0.68 \\
 & 2 & \textbf{0.74} & 0.70 & 0.70 & 0.73 & 0.66 & 0.59 \\
\midrule
\multirow{2}{*}{\textbf{AUC}}
 & 1 & \textbf{0.93} & 0.92 & 0.88 & 0.88 & 0.85 & 0.74 \\
 & 2 & \textbf{0.82} & 0.77 & 0.77 & 0.79 & 0.72 & 0.64 \\
\midrule
\multirow{2}{*}{\textbf{TPR}}
 & 1 & 0.84 & \textbf{0.86} & 0.81 & 0.77 & 0.66 & 0.68 \\
 & 2 & 0.66 & \textbf{0.82} & 0.74 & 0.71 & 0.59 & 0.44 \\
\midrule
\multirow{2}{*}{\textbf{TNR}}
 & 1 & 0.88 & 0.84 & 0.83 & \textbf{0.90} & 0.88 & 0.68 \\
 & 2 & \textbf{0.82} & 0.59 & 0.66 & 0.75 & 0.73 & 0.75 \\
\bottomrule
\end{tabular}
\caption{Detection performance (on the full private set) with a comparison of the two tasks, the top teams and the baseline across four metrics: balanced accuracy (BAC), area under the curve (AUC), true positive rate (TPR), and true negative rate (TNR)}.
\label{tab:task_comparison_transposed_multirow}
\end{table}
\textbf{Post-Processing:}
To test robustness of detectors to post-processing, in Task 2 we used 14 post-processing techniques that are commonly performed on videos found on the Internet. These techniques included resolution manipulation with upscaling and downscaling (aspect ratio fixed; upscale = height x 1.75; downscale  = height x .75), and common compression algorithms: H.264, H.265, and AV1. We also included other techniques such as adding a border, color correction, frame interpolation, gaussian blur, motion blur, noise, speed adjustment, and text overlaying. Finally, we included one laundering technique ("real camera") where we recorded the video playing on a screen with another camera. Each technique along with the description is listed in Table~\ref{tab:post_processing_technique}. Videos from Task 1 were uniformly sampled from each source, resulting in a total of 78 synthetic videos and 84 real videos. Each post-processing technique was applied to every video, for a total of 1092 synthetic videos and 1176 real videos. The partitioning of Task 2 videos to either public or private datasets corresponded to the partitioning of their original videos as used in Task 1. Note that the augmentation did not result in noticeable quality degradation or the presence of artifacts (see Figure \ref{fig:video_augs} ) thus qualifying as reasonable processing that one could do to videos before sharing. 


\newcommand{\codesource}[1]{%
  {\ttfamily\hyphenchar\font=`\- #1}%
}

\renewcommand{\tabularxcolumn}[1]{>{\raggedright\arraybackslash}m{#1}}

\begin{table}[tbh!]
\centering
\small
\setlength{\tabcolsep}{6pt}
\renewcommand{\arraystretch}{1.15}
\begin{tabularx}{\linewidth}{X r r r r r r}
\toprule
\textbf{Source} & \textbf{$\mu$} & \textbf{GR} & \textbf{IS} & \textbf{DA} & \textbf{LE} & \textbf{TR} \\
\midrule
\codesource{vsp-trail-cam} & 0.97 & 0.98 & \textbf{1.00} & 0.94 & 0.93 & 0.99 \\
\codesource{vsp-drone} & 0.91 & \textbf{0.99} & \textbf{0.99} & 0.84 & 0.84 & 0.92 \\
\codesource{social-media-dance} & 0.91 & \textbf{0.98} & 0.92 & 0.90 & 0.87 & 0.90 \\
\codesource{drone} & 0.91 & 0.95 & \textbf{0.97} & 0.83 & 0.83 & \textbf{0.97} \\
\codesource{social-media-text-overlays} & 0.91 & \textbf{0.97} & 0.90 & 0.92 & 0.86 & 0.88 \\
\codesource{lecture} & 0.90 & 0.96 & \textbf{0.98} & 0.94 & 0.88 & 0.80 \\
\codesource{vlog} & 0.90 & 0.95 & \textbf{0.99} & 0.92 & 0.86 & 0.90 \\
\codesource{social-media-composite-video} & 0.90 & \textbf{0.97} & \textbf{0.97} & 0.89 & 0.88 & 0.89 \\
\codesource{documentary} & 0.90 & \textbf{0.97} & 0.95 & 0.88 & 0.86 & 0.89 \\
\codesource{time-lapse} & 0.90 & 0.95 & \textbf{0.96} & 0.91 & 0.88 & 0.84 \\
\midrule
\texttt{Public Avg} & 0.89 & 0.92 & \textbf{0.94} & 0.87 & 0.87 & 0.86 \\
\midrule
\codesource{academic-selfie} & 0.90 & \textbf{1.00} & 0.97 & 0.85 & 0.87 & 0.91 \\
\codesource{vsp-selfie} & 0.90 & \textbf{1.00} & 0.96 & 0.85 & 0.84 & 0.88 \\
\codesource{vsp-travel} & 0.90 & \textbf{0.99} & 0.98 & 0.83 & 0.88 & 0.88 \\
\codesource{vsp-production} & 0.90 & \textbf{0.98} & \textbf{0.98} & 0.83 & 0.88 & 0.87 \\
\codesource{social-media-phone-camera} & 0.89 & \textbf{0.98} & 0.96 & 0.88 & 0.86 & 0.91 \\
\codesource{internet-archive-sports} & 0.88 & 0.87 & \textbf{0.98} & 0.94 & 0.88 & 0.83 \\
\codesource{internet-archive-pre-digital-footage} & 0.85 & 0.85 & \textbf{0.99} & 0.92 & 0.87 & 0.61 \\
\codesource{sports} & 0.85 & 0.87 & \textbf{0.91} & 0.81 & 0.86 & 0.84 \\
\codesource{vsp-tv-talk-show} & 0.80 & 0.85 & \textbf{0.92} & 0.82 & 0.84 & 0.90 \\
\codesource{dashcam} & 0.78 & 0.88 & 0.49 & \textbf{0.95} & 0.90 & 0.68 \\
\midrule
\texttt{Private Avg} & 0.89 & \textbf{0.95} & 0.90 & 0.88 & 0.87 & 0.85 \\
\bottomrule
\end{tabularx}
\caption{AUC conditioned on the real source. Only the top five teams and their means are shown for brevity.}
\label{tab:auc_by_real}
\end{table}


\renewcommand{\tabularxcolumn}[1]{>{\raggedright\arraybackslash}m{#1}}

\begin{table}[tbh!]
\centering
\small
\setlength{\tabcolsep}{6pt}
\renewcommand{\arraystretch}{1.15}
\begin{tabularx}{\linewidth}{X r r r r r r}
\toprule
\textbf{Source} & \textbf{$\mu$} & \textbf{GR} & \textbf{IS} & \textbf{DA} & \textbf{LE} & \textbf{TR} \\
\midrule
\codesource{veo-2} & 0.97 & \textbf{1.00} & 0.95 & 0.99 & 0.99 & 0.94 \\
\codesource{wan-2\_1-i2v-480p} & 0.97 & \textbf{1.00} & 0.99 & 0.98 & 0.97 & 0.93 \\
\codesource{pixverse-v4\_5} & 0.96 & \textbf{0.99} & 0.95 & 0.97 & 0.98 & 0.89 \\
\codesource{kling-v2\_0} & 0.95 & \textbf{0.99} & 0.95 & 0.91 & 0.95 & 0.94 \\
\codesource{hunyuan} & 0.94 & \textbf{0.99} & 0.92 & 0.95 & 0.95 & 0.90 \\
\codesource{framepack} & 0.85 & \textbf{0.93} & 0.81 & 0.92 & 0.92 & 0.77 \\
\midrule
\texttt{Public Avg} & 0.94 & \textbf{0.98} & 0.93 & 0.95 & 0.96 & 0.89 \\
\midrule
\codesource{runway-gen4-turbo} & 0.98 & \textbf{0.99} & 0.98 & 0.99 & 0.99 & 0.95 \\
\codesource{cosmos-drive-dreams} & 0.92 & \textbf{0.94} & 0.92 & 0.88 & 0.86 & 0.88 \\
\codesource{vidu-2.0} & 0.88 & 0.84 & \textbf{0.90} & 0.81 & 0.77 & 0.81 \\
\codesource{video-01} & 0.81 & 0.87 & \textbf{0.96} & 0.62 & 0.62 & 0.70 \\
\codesource{hunyuan-avatar} & 0.79 & 0.92 & \textbf{0.95} & 0.62 & 0.69 & 0.61 \\
\codesource{seedance-1-pro} & 0.77 & 0.83 & \textbf{0.87} & 0.78 & 0.69 & 0.90 \\
\codesource{physicalai} & 0.75 & 0.87 & 0.77 & \textbf{0.98} & 0.72 & 0.82 \\
\midrule
\texttt{Private Avg} & 0.84 & 0.89 & \textbf{0.91} & 0.81 & 0.80 & 0.81 \\
\bottomrule
\end{tabularx}
\caption{AUC conditioned on the generator. Only the top five teams and their means are shown for brevity.}
\label{tab:auc_by_gen}

\end{table}

\section{Results}
Here we present an overview of the results from the SAFE Video Detection Challenge\footnote{Competition details: \url{https://safe-video-2025.dsri.org} }. The challenge ran for 90 days and attracted participation from over 12 teams worldwide, resulting in more than 600 model submissions across both tasks. To contextualize leaderboard results, Table~\ref{tab:team_mapping} maps the Hugging Face submission handles used during evaluation to the corresponding institutions for top performing teams that presented results at the workshop. Due to space limitations, result tables show only the top 5 teams (abbreviated by the first two letters). The charts show additional teams. Also note that we have incomplete details of participants' approaches since they were not required to share any specific details (see Table \ref{tab:team_methods}). For a baseline (BA), we used a synthetic video detector from \cite{vahdati2024} that has been trained on several older synthetic video models such Stable Video Diffusion and COG.

From the competition results, we observe several important takeaways. Table \ref{tab:task_comparison_transposed_multirow} shows performance across the top teams, as well as several metrics across the two tasks.

\textbf{Strong Performance of Unmodified Synthetic Videos:} Performance on detection of unmodified synthetic video (Task 1) was notably strong. The top teams, \textbf{GR}IP-UNINA, \textbf{IS}PL-Realynx, \textbf{DA}SH, and \textbf{LE}mma1727 achieved an AUC of 0.93, 0.92, 0.88, and 0.88, respectively, substantially outperforming the baseline model (0.74).  Generalization of submissions on both private and public splits was surprising given that the participants had no knowledge of synthetic and real sources in either split.

\textbf{Detectors are Not Robust to Post-processed Content:} Across all teams, performance was consistently lower on Task 2 (detection after post-processing). This suggests that even the best techniques are still not robust to common operations that videos undergo when shared across social media and the Internet.

\textbf{Pre-trained Image Backbones and Autoencoding for Training Good Detectors:} Top performers shared several important aspects in their model training and architecture. First, all top performers utilized a strong backbone (DINOv2) specifically designed for general purpose computer vision tasks (object detection, segmentation, etc.) and pre-trained on large amount of real images. Second, most performers incorporated autoencoding augmentation during their training. A real image or video is reconstructed using the VAE creating a synthetic example perfectly matched to a real one w.r.t. content. Table \ref{tab:team_methods} shows different aspects of participants' approaches \footnote{Not all teams shared approach details with the organizers.} 

\textbf{Surprising Generalizability to Novel Generators:} Table \ref{tab:team_methods} shows that the top performing teams had little or no overlap between the challenge and training data while still achieving very high performance. This suggests a significant degree of transferability to unknown detectors. For example, \textbf{DA}SH only used a single model (WAN2.2) for training while achieving good performance across the 12 other models in our test set. This suggests that the architecture and the training setup is just as important as having a wide coverage of a diverse array of generators.
\begin{figure*}[tbh!]
\centering
\includegraphics[width=.95\textwidth]{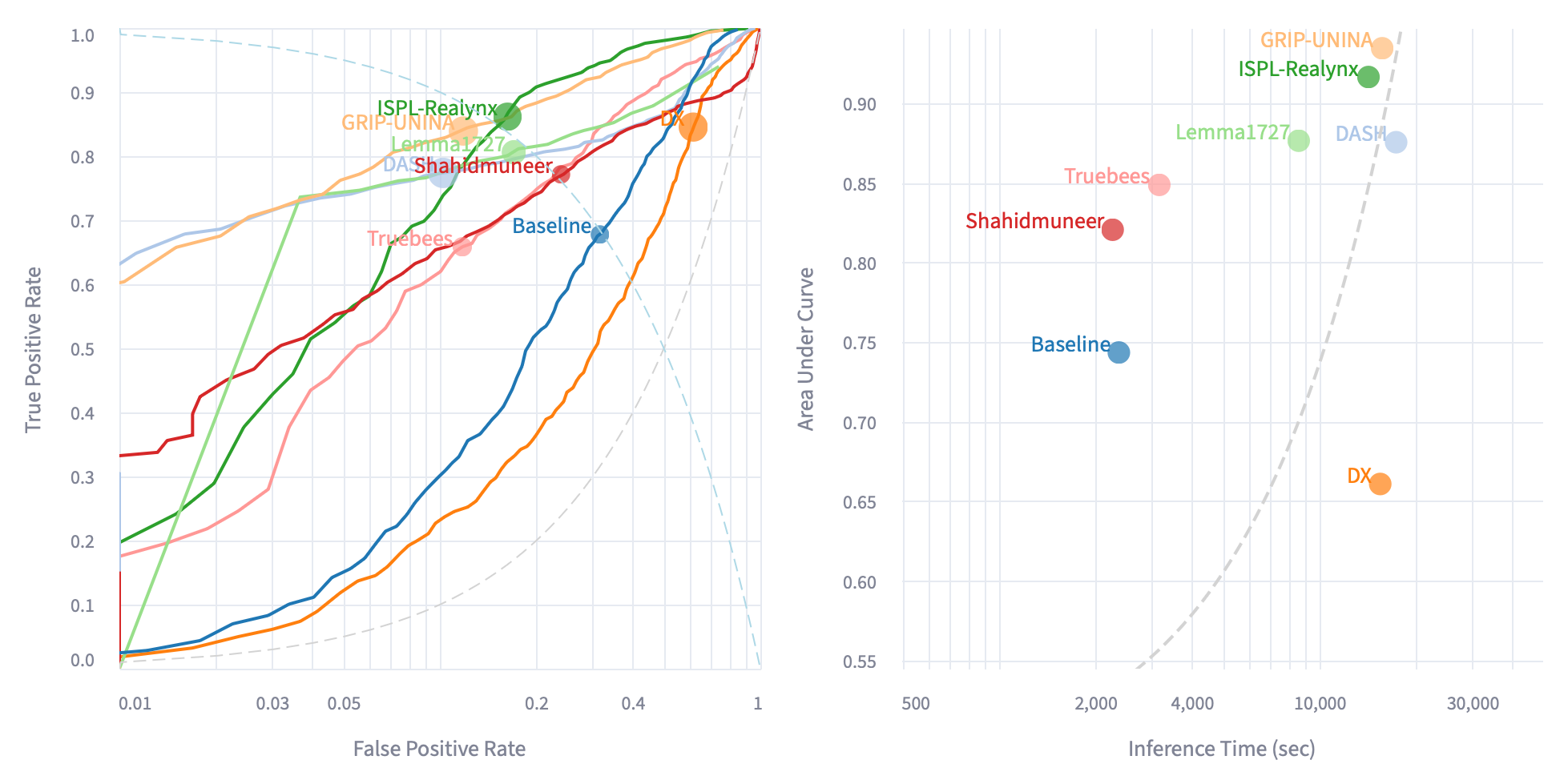}
\caption{Task 1 Results: [Left] ROC Curve for all teams (with FPR plotted on a log scale. [Right] AUC vs Inference time (on a log scale)}
\label{fig:task1_roc}
\end{figure*}

\begin{figure}[tbh!]
\centering
\includegraphics[width=.49\textwidth]{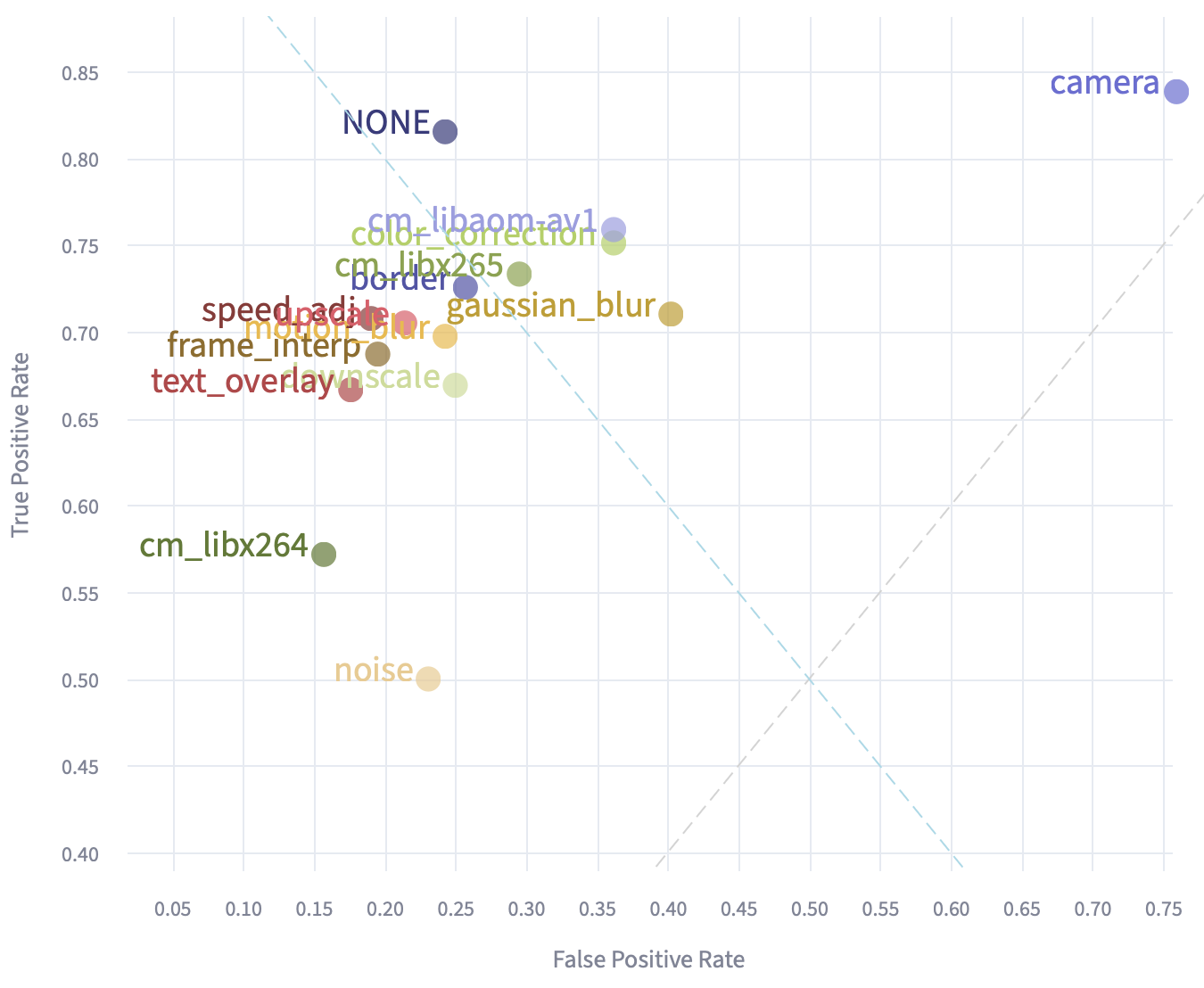}
\caption{True vs False Positive Rates for each augmentation type in Task 2 for the best GR are shown. Details are in Table \ref{tab:post_processing_technique}.}
\label{fig:task2_scatter}
\end{figure}


\renewcommand{\tabularxcolumn}[1]{>{\raggedright\arraybackslash}m{#1}}

\begin{table}[tbh!]
\centering
\small
\setlength{\tabcolsep}{6pt}
\renewcommand{\arraystretch}{1.15}
\begin{tabularx}{\linewidth}{X r r r r r r}
\toprule
\textbf{Augmentation} & \textbf{$\mu$} & \textbf{GR} & \textbf{DA} & \textbf{IS} & \textbf{LE} & \textbf{TR} \\
\midrule
\codesource{none} & 0.88 & \textbf{0.93} & 0.87 & 0.90 & 0.84 & 0.84 \\
\codesource{speed-adjustment} & 0.83 & 0.87 & 0.84 & \textbf{0.89} & 0.82 & 0.75 \\
\codesource{upscale} & 0.83 & \textbf{0.89} & 0.82 & 0.83 & 0.80 & 0.81 \\
\codesource{text-overlay} & 0.83 & 0.86 & 0.83 & \textbf{0.89} & 0.82 & 0.75 \\
\codesource{border} & 0.82 & \textbf{0.88} & 0.80 & 0.85 & 0.79 & 0.76 \\
\codesource{frame-interpolation} & 0.82 & 0.85 & 0.83 & \textbf{0.86} & 0.81 & 0.73 \\
\codesource{motion-blur} & 0.81 & 0.84 & 0.82 & \textbf{0.86} & 0.80 & 0.72 \\
\codesource{compression-libx265} & 0.79 & \textbf{0.87} & 0.79 & 0.83 & 0.74 & 0.71 \\
\codesource{downscale} & 0.79 & 0.81 & \textbf{0.82} & 0.81 & 0.76 & 0.72 \\
\codesource{color-correction} & 0.78 & \textbf{0.85} & 0.79 & 0.77 & 0.78 & 0.69 \\
\codesource{compression-libaom-av1} & 0.77 & \textbf{0.85} & 0.76 & 0.74 & 0.75 & 0.75 \\
\codesource{compression-libx264} & 0.77 & 0.75 & \textbf{0.81} & 0.75 & \textbf{0.81} & 0.70 \\
\codesource{gaussian-blur} & 0.74 & 0.73 & 0.75 & \textbf{0.79} & 0.76 & 0.67 \\
\codesource{noise} & 0.72 & \textbf{0.83} & 0.75 & 0.63 & 0.68 & 0.71 \\
\codesource{camera} & 0.62 & 0.66 & \textbf{0.67} & 0.51 & 0.63 & 0.63 \\
\bottomrule
\end{tabularx}
\caption{Task 2 Results: AUC conditioned on augmentation (for the top team and their means) are shown. The rows are sorted by decreasing AUC. The "none" row corresponds to unmodified videos.}
\label{tab:auc_aug}
\end{table}

\renewcommand{\tabularxcolumn}[1]{>{\raggedright\arraybackslash}m{#1}}

\begin{table*}[tbh!]
\centering
\small
\setlength{\tabcolsep}{6pt}
\renewcommand{\arraystretch}{1.15}
\begin{tabularx}{\linewidth}{l X X X X X}
\toprule
\textbf{Team} & \textbf{Synthetic} & \textbf{Real} & \textbf{Overlap} & \textbf{Augmentation} & \textbf{Architecture} \\
\midrule
\textbf{BA \cite{vahdati2024}} & Video Crafter, Luma, SVD, CogVideo & VideoACID, Moments in Time & None & .. & MISLNet\\ 
\textbf{GR \cite{corvi2025seeing}} & Pyramid Flow, COG Video X, Open Sora, Allegro & .. & No Synthetic & autoencoding, wavelet replacement & Ensemble w. DINOv2/3 backbone, Frame Average \\
\textbf{IS} & 45+ different generators & variety of video frames, images & Limited
 & Autoencoding, Other & EfficientNet, Frame Average  \\
\textbf{DA/LE} & Wan2.2  & 2k HQ Youtube Vids & Only WAN2.2 & Autoencoding, Real Frame to Video & DINOv2 Backbone + Transformer over Frames \\
\bottomrule
\end{tabularx}
\caption{Overview of training strategies, data augmentation, and architectures by team. Note that this table is incomplete since teams were not required to share the details of their approach. (The DASH and LEmma teams had very similar approaches).}
\label{tab:team_methods}
\end{table*}

\subsection{Task 1 Discussion}
Tables \ref{tab:auc_by_real} and \ref{tab:auc_by_gen} show AUC conditioned on real sources and video generators, respectively. During the competition, participants had access to these fine-grained performance metrics but only on the public split of the data (the top half of the tables). However, the names of the sources remained anonymous. Performance on the private set remained hidden until the end of the competition. Even under these restrictions, the results did not show any large performance gaps between private and public sources.

In Table \ref{tab:auc_by_gen} showing performance conditioned on the video generator,  the mean AUCs ($\mu$) of top teams was lower on average between public (.94) and private (.84). The generators (hunyan-avatar, seedance-1-pro, physicalai) had the lowest performance in the mid to upper 70's compared to scores in the 90's for most generators in the public split. This is likely due to hunyan-avatar being an audio+text+image to video model and seedance-1-pro being a newer generator. The physicalai data is a CGI-like physical motion generator model that is somewhat different from standard I2V models. For details on different synthetic video sources see Table \ref{tab:synthetic_video_sources}.

In Table \ref{tab:auc_by_real}, the mean AUCs conditioned on real sources have the same averages for private and public splits among the top teams. The only outliers appear to be the "dashcam" and the "tv-talk-show" source (with .78 and .80 AUCs). The former is both semantically and forensically different than other the sources which could explain performance drops.

Figure \ref{fig:task1_roc} shows the ROC curves for the best submission. Note the x-axis (False Alarm Rate) is on the log scale. A large fraction of submissions fell on the equal error rate curve suggesting balanced true and false positive rates. However, none of the detectors can operate in very low false alarm regimes (<.01). Therefore, while the overall performance is promising, we are still far from being able to handle large (i.e. Internet) scale detection of synthetic videos without being flooded with false alarms. The right panel shows AUC vs inference time (on a log scale). Since all submissions used the same underlying hardware, this figure reflects realistic inference times. The main trend is that a longer inference time (likely implying a model with more parameters) improves detection. However, we see, for example, that the Lemma and DX team achieved the same AUC with the Lemma being significantly faster.
\subsection{Task 2 Discussion}
Recall that in Task 2, we applied a set of augmentation and post-processing operations to the subset of video data from Task 1 to simulate common operations that occur when sharing videos on the Internet. Table \ref{tab:auc_aug} shows the AUC conditioned on the augmentation type for the top submissions (sorted by the mean $\mu$ performance). The main observation is that all augmentations reduced the detection accuracy by 5 to 25 AUC points when compared to unaltered video (the "none" category in the table \footnote{For real video, we do not know the complete history of processing. }). 

In Figure \ref{fig:video_augs} we also see that visually most of the augmentations applied are very mild in terms of reducing visual quality or adding semantic artifacts. Of note is the significant drop in performance due to video re-compression with "libx264 and libx265, and av1" reducing performance from .88 to .77, for libx264 and av1 and to .79 for libx265. This is likely due to the additional corruption of forensic traces present in the video. Gaussian blurring and white noise augmentations resulted in average AUC scores of .74 and .72, respectively, likely for the same reason. The last and most difficult augmentation was the laundering "camera" process where we recorded a video playing on a screen with another camera as a digital-to-analog-to-digital operation thus the drop to .62 from .88 is not surprising.

Figure \ref{fig:task2_scatter} shows a scatter plot of the average true and false positive rates color coded by the augmentation type. We observed that different augmentations resulted in different error types. For example, some operations such as "noise" only degraded true positive rates (which makes sense since real videos have naturally occurring noise due to the physical process of imaging the real world while synthetic videos do not). Some degraded both error types such as gaussian blur. Surprisingly, the laundering “camera” augmentation increased false alarm rates, indicating that at the detectors’ operating points, laundered real videos shift closer in score to synthetic videos, contrary to our expectation that the opposite effect would occur. 

\subsection{Discussion of Participants' Approaches}
Table \ref{tab:team_methods} contains partial details on submission approaches including training data used, augmentation techniques applied, and architecture of detection models. Participants were not required to provide all details of their approach and, by design, the competition did not have access to the submission code or models; therefore, the table covers only a subset of teams. As mentioned earlier, we observed interesting patterns across the top submissions. A few submissions such as {\bf GR}IP-UNINA and {\bf DA}SH had almost no overlap with competition in their training while achieving very good detection accuracy suggesting strong transferability between different synthetic video generators. The use of autoencoding on real video images to generate synthetic training examples emerged as an effective augmentation strategy for improving model performance. Lastly, starting from a large scale general purpose computer vision backbone such DinoV2 that had been trained on large amounts of real images was considered an important component for training synthetic image or video detectors.

\section{Acknowledgments}
This work and the competition are sponsored by ULRI Digital Safety Research Institute.

\bibliographystyle{elsarticle-num} 
\bibliography{references-compact}
\appendix

\section{Additional Competition Details}
Additional details about the competition are provided below.
\begin{figure*}[!t]
\centering
\includegraphics[trim = 2cm 2cm 2cm 2cm,
  clip,
  width=.95\linewidth]{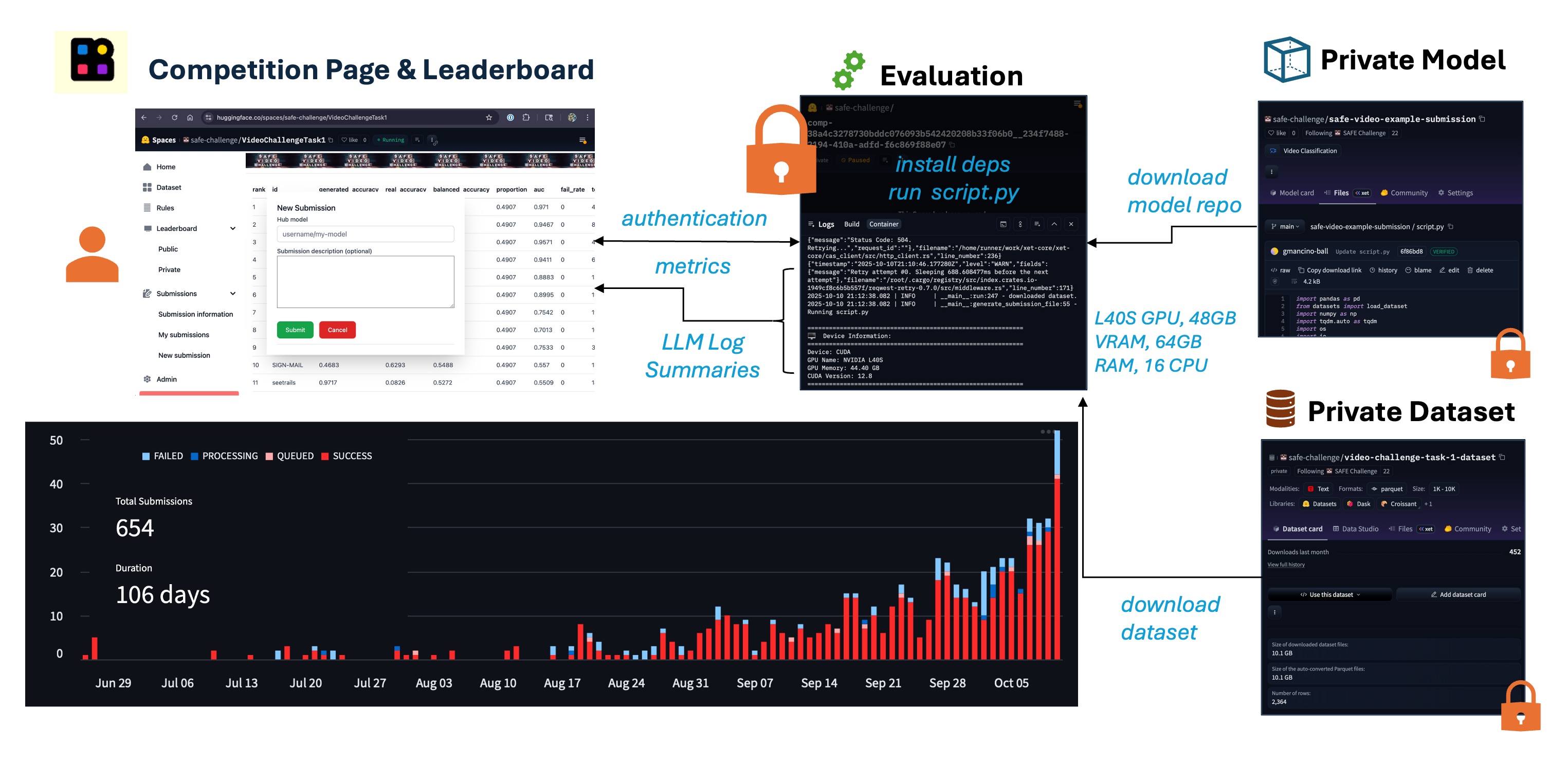}
\caption{Competition Setup: participants log in with their Hugging Face credentials and provide the submission repo. In the backed, our platform pulls the model, installs the requirements, pulls the private dataset and runs the evaluation without any network access. The bottom left shows the volume of our competition submissions, which spiked a month before the end of the challenge.}
\label{fig:comp_setup}
\end{figure*}



\end{document}